\documentclass[10pt, a4paper]{article}
\usepackage{lrec2022} % this is the new LREC2022 Style
\usepackage{multibib}
\newcites{languageresource}{Language Resources}
\usepackage{graphicx}
\usepackage{tabularx}
\usepackage{soul}
\usepackage{mathtools}
\usepackage{amssymb}
\usepackage{mathrsfs}
\usepackage{amsmath}
\usepackage{titlesec}
\titleformat{\section}{\normalfont\large\bfseries\center}{\thesection.}{1em}{}
\titleformat{\subsection}{\normalfont\SmallTitleFont\bfseries\raggedright}{\thesubsection.}{1em}{}
\titleformat{\subsubsection}{\normalfont\normalsize\bfseries\raggedright}{\thesubsubsection.}{1em}{}
\renewcommand\thesection{\arabic{section}}
\renewcommand\thesubsection{\thesection.\arabic{subsection}}
\renewcommand\thesubsubsection{\thesubsection.\arabic{subsubsection}}
\usepackage{epstopdf}
\usepackage[utf8]{inputenc}

\usepackage{hyperref}
\usepackage{xstring}

\usepackage{wasysym} % long s

\usepackage{color}
\usepackage{xcolor}
\usepackage{xspace}
\usepackage{booktabs}

\newcommand{\roberta}{RoBERTa\xspace}
\newcommand{\bert}{BERT\xspace}
\newcommand{\camembert}{CamemBERT\xspace}
\newcommand{\dalembert}{D'AlemBERT\xspace}
\newcommand{\freemmax}{\textsc{FreEM}\textsubscript{\emph{max}}\xspace}
\newcommand{\freemlpm}{\textsc{FreEM}\textsubscript{\emph{LPM}}\xspace}
\newcommand{\pieextended}{Pie Extended\xspace}

\title{From \textsc{FreEM} to D’AlemBERT: a Large Corpus and a Language Model for Early Modern French}

\name{Simon Gabay$^{3}$, Pedro Ortiz Suarez$^{1,2}$, Alexandre Bartz$^{2}$, Alix Chagué$^{1}$\\
    \large\textbf{Rachel Bawden$^{1}$, Philippe Gambette$^{4}$, Benoît Sagot$^{1}$}}

\address{Inria$^1$, Sorbonne Universit\'e$^2$, Universit\'e de Gen\`eve$^3$,  LIGM, Universit\'e Gustage Eiffel, CNRS$^4$\\
2 rue Simone Iff, 75012 Paris (France)$^1$, 21 rue de l’École de médecine, 75006 Paris (France)$^2$,  \\ rue du Général-Dufour 24, 1211 Genève (Switzerland)$^3$, \\
5 boulevard Descartes, F-77454 Champs-sur-Marne (France)$^4$ \\
\{pedro.ortiz, alix.chague, benoit.sagot, rachel.bawden\}@inria.fr$^1$,
alexandre.bartz@sorbonne-universite.fr$^2$, \\ simon.gabay@unige.ch$^3$,  philippe.gambette@univ-eiffel.fr$^4$}

\abstract{
    Language models for historical states of language are becoming increasingly important to allow the optimal digitisation and analysis of old textual sources. Because these historical states are at the same time more complex to process and more scarce in the corpora available, specific efforts are necessary to train natural language processing (NLP) tools adapted to the data. In this paper, we present our efforts to develop NLP tools for Early Modern French (historical French from the 16\textsuperscript{th} to the 18\textsuperscript{th} centuries). We present the \freemmax corpus of Early Modern French and D'AlemBERT, a RoBERTa-based language model trained on \freemmax. We evaluate the usefulness of D'AlemBERT by fine-tuning it on a part-of-speech tagging task, outperforming previous work on the test set. Importantly, we find evidence for the transfer learning capacity of the language model, since its performance on lesser-resourced time periods appears to have been boosted by the more resourced ones. We release D'AlemBERT and the open-sourced subpart of the \freemmax corpus.
    \\ \newline  }

\begin{document}

\maketitleabstract

\section{Introduction}

With the rise of digital humanities, it is becoming increasingly important to develop high quality tools to automatically process old states of languages. Libraries, archives and museums, among others, are digitising large numbers of historical sources, from which high quality data must be extracted for further study by specialists of human sciences following new approaches such as ``distant reading'' \cite{moretti_distant_2013}. Many (sub)tasks such as automatic OCR post-correction \cite{rijhwani_lexically_2021} and linguistic annotation \cite{camps_corpus_2020} benefit from pretrained language models to improve their accuracy, and this is what motivated us to develop a BERT-like \cite{devlin-etal-2019-bert} contextualised language model for Early Modern French.

Languages evolve over time on many different levels: from one century to another, we see variations in spelling, syntax, the lexicon etc. However this variation is not uniform: it tends, at least for ``literate scriptors'' (literature, journalism, law, etc.), to converge towards a single norm over time, and this has especially been the case for French because of the prominent role of the \textit{Académie française} and the \emph{remarqueurs} \cite{ayres-bennett_remarques_2011}. The result of this convergence is, for instance, that spelling and word order within sentences have become more strict, where they were less so in the past. From a computational perspective, historical states of language are therefore not only different from the contemporary state, but, from a computational perspective, are also more complex because they do not follow a strict and explicit norm. In French, this explicit norm  appeared in the 17\textsuperscript{th}~c. and was slowly integrated throughout the 18\textsuperscript{th}~c.

On top of this first linguistic problem, a second issue appears: because the production of textual sources has continued to grow exponentially, it is easier to collect a corpus for contemporary French  than for the 19\textsuperscript{th}\,c. French, which is itself easier than for the 18\textsuperscript{th}\,c. French, etc. The further we go back in time, the more scarce resources are, which creates the following paradox: we have more data when the language is homogeneous and simple for the computer to process, and less when it is heterogeneous and harder to process.

The following paper will address the development of \dalembert a neural language model in a complex setting, defined here as the state of language with scarce heterogeneous resources. We will also present \freemmax the data used to train the model, discuss its conception, and evaluate its efficiency with a classical natural language processing (NLP) task, part-of-speech (POS) tagging, crucial for corpus linguistics and the digital humanities. We release both the \dalembert model and a subset of the \freemmax dataset that we were allowed to open source by the original authors.

\section{Related Work}

Large datasets for historical states of languages or  extinct languages do exist. The \emph{Corpus Middelnederlands} for Medieval Dutch \citelanguageresource{Reenen_corpus_1998} and the \emph{Base Geste} for Medieval French \citelanguageresource{camps_geste_2016} are freely available online, encoded in TEI. It is also the case for other corpora for later states of language, such as the \textit{Reference corpus of historical Slovene}, covering approximately three centuries of Slovene (1584--1899)  \citelanguageresource{erjavec_reference_2015}, and the ``corpus noyau'' of \emph{Presto} \citelanguageresource{presto_corpus_2018}. This last corpus, in its extended version, uses other French corpora such as \textit{Espistemon} for Renaissance French \citelanguageresource{demonet_epistemon_1998} and the University of Chicago's \emph{American and French Research on the Treasury of the French Language} (ARTFL) \citelanguageresource{morrissey_artfl_1981}; or like \textsc{Frantext} \citelanguageresource{atilf_frantext_1998}, which is a generalist French corpus, covering the different states of the French language between the 11\textsuperscript{th} and the 21\textsuperscript{st} century. Although most of these text collections are free, the two biggest ones, \textsc{Frantext} and ARTFL, are not freely available or open-sourced.

Concerning language modelling in French, two main models are available for contemporary French, \camembert \cite{martin-etal-2020-camembert} and FlauBERT \cite{le-etal-2020-flaubert}. \camembert was trained on a freely available, automatically web-crawled corpus called OSCAR \cite{ortiz-suarez-etal-2019-asynchronous,ortiz-suarez-etal-2020-monolingual} while FlauBERT was trained on a mix of web-crawled data and manually curated (partly non freely available) contemporary French corpora. Neither of these models was explicitly pre-trained for historical French.\footnote{Note however that texts in Old, Middle and Modern French do exist in the internet, and might have found their way to the training corpus of these two models. This is especially the case for Modern French texts, which automatic language classification tools can easily classify as Contemporary French.} However efficient language models have been trained for less-resourced or extinct Languages such as Latin \cite{Bamman_latinBErt_2020}, following the approach of \newcite{martin-etal-2020-camembert} for training language models with less data than was previously thought. There have also been some recent projects that specifically target Early Modern French such as that of \pieextended \cite{clerice-2020-pie} that uses the hierarchical encoding architecture originally proposed by \newcite{manjavacas-etal-2019-improving} which itself is constructed by stacking multiple Bi-LSTM-CRFs. \newcite{clerice-2020-pie} distributes pre-trained models for POS tagging and lemmatisation.

\section{Corpora}

\begin{table*}[!htp]
    \centering\small
    \begin{tabular}{@{}p{0.3\linewidth}p{0.3\linewidth}p{0.3\linewidth}@{}}
        \toprule
        Source & Normalised & Translation \\
        \midrule
        Surquoy, SIRE, s’il plai\longs{}t à vo\longs{}tre Maie\longs{}té de \longs{}e \longs{}ouuenir des mi\longs{}eres de \longs{}on E\longs{}tat, dõt au moins ell’a tiré cét aduantage, qu’en vne grande ieune\longs{}se ell’a acquis vne grande experi\~ece, elle verra que tous les mal-heurs de sõ bas âge ont pris leur commencement en \longs{}emblables occa\longs{}ions; &
        \textit{Sur quoi, SIRE, s’il plaît à votre Majesté de se souvenir des misères de son état dont au moins elle a tiré cet avantage, qu’en une grande jeunesse elle a acquis une grande expérience, elle verra que tous les malheurs de son bas âge ont pris leur commencement en semblables occasions~;} &
        \textcolor{gray}{``Whereupon, SIR, if it pleases your Majesty to remember the miseries of her state, from which at least she has derived this advantage, that in great youth she has acquired great experience, she will see that all the misfortunes of her early life took their beginning on similar occasions;''}\\
        \bottomrule
    \end{tabular}
    \caption{\label{tab:norm_examples}Example of normalisation taken from the \textit{Lettres} of \protect\newcite{guez_de_balzac_lettres_1624}.}
\end{table*}

For the past few years, we have been involved in the development of linguistic resources for Early Modern French. The initiative, called \textsc{FreEM} (which stands for \textit{FREnch Early Modern}), aims to collect the corpora required for various NLP tasks such as lemmatisation, POS tagging, linguistic normalisation and named entity recognition. Two of these corpora are introduced here: \freemmax (see Section~\ref{freem_max}) and \freemlpm (see Section~\ref{freem_lpm}).

\subsection{Early Modern French}\label{def:early}

Experiments are based on data of which the core comprises Early Modern French literary texts. We loosely define Early Modern French as a state of language following Middle French in 1500---following here the \textit{terminus ad quem} used by the \textit{Dictionnaire de Moyen Français} \cite{martin_dictionnaire_2020}---and ending with the French Revolution in 1789. It therefore encompasses three centuries (16\textsuperscript{th}, 17\textsuperscript{th} and 18\textsuperscript{th}\,c.), or two linguistic periods: the \textit{français préclassique} or ``preclassical French'', 1500--1630 and the \textit{français classique} or ``classical French'', 1630--1689; both periodisations are currently used in French linguistics (\textit{e.g.}~by \newcite{vachon_changement_2010} and \newcite{amatuzzi_ameliorer_2019}).

A typical example of Early Modern French, taken from ~\newcite{guez_de_balzac_lettres_1624}, is given in Table~\ref{tab:norm_examples}. We note here the presence of several phenomena that have now disappeared in contemporary French, such as the presence of abbreviations (\textit{dõt}$\to$\textit{dont}), the long \textit{s} (\textit{\longs{}}, see\,\textit{mi\longs{}eres}), the use of \textit{v} instead of \textit{u} (\textit{vne} for \textit{une}), the conservation of etymological letters (\textit{vo\longs{}tre}$<$Latin~\textit{vŏster} rather than \textit{votre}) and calligraphic letters (\textit{-y} in \textit{Surquoy}), the absence of welding  (\textit{\mbox{mal-heurs}} and not \textit{malheurs}) and the opposite (\textit{Surquoy} and not \textit{Sur quoi}).

For NLP tasks, which process raw sequences, such differences with respect to contemporary French are not trivial, and they prevent the processing of historical texts with tools trained on recent sources.

\subsection{\texorpdfstring{\freemmax}{FREEM max}}\label{freem_max}

Usable historical documents are difficult to find because, as previously mentioned, they are more rare than contemporary ones; editors tend to normalise the language (\textit{i.e.}~use the spelling conventions of contemporary French, see~\cite{gabay_pourquoi_2014}), transcriptions are not (always) distributed in a digital format. \freemmax \citelanguageresource{FREEMmax} is an attempt to solve this problem, and the aim of this dataset is to group together the largest number or texts possible written in Early Modern French. 
The texts have a variety of sources, which can be grouped into three main types:
\begin{itemize}
    \item Two institutional datasets have been used and are non open-sourced:
    \begin{itemize}
        \item \textsc{Frantext} \textit{intégral} \citelanguageresource{atilf_frantext_1998}, the biggest database of French texts (only the texts between 1500 and 1800), a very small portion of which is open access: \textsc{Frantext} \textit{Démonstration} \citelanguageresource{atilf_frantext_1998-1};
        \item \textit{Electronic Enlightenment} \citelanguageresource{bodleian_libraries_electronic_2008}, an online collection of edited correspondences of the Early Modern period;
    \end{itemize}
    \item Several come from research projects distributing transcriptions online:
    \begin{itemize}
        \item The \textit{Antonomaz project},  French \textit{mazarinades} (\url{https://cahier.hypotheses.org/antonomaz});
        \item The II.B section (in French) of the \textit{Actis Pacis Westphalicae}, diplomatic letters for the Peace of Westphalia (\url{http://kaskade.dwds.de/dstar/apwcf/});
        \item The Bibliothèques virtuelles humanistes, 16\textsuperscript{th}\,c.~French literature (\url{http://www.bvh.univ-tours.fr});
        \item The \textit{Corpus électronique de la première modernité}, 17\textsuperscript{th}\,c.~French literature (\url{http://www.cepm.paris-sorbonne.fr})
        \item The \textit{Condé} project, \textit{coutumiers normands} (\url{https://conde.hypotheses.org})
        \item The Corpus Descartes, works of René Descartes (\url{https://www.unicaen.fr/puc/sources/prodescartes/});
        \item The \textit{Bibliothèque dramatique} of the CELLF, 17\textsuperscript{th}\,c.~French plays (\url{http://bibdramatique.huma-num.fr});
        \item The \textit{Fabula numerica} project, French fables (\url{https://obvil.sorbonne-universite.fr/projets/fabula-numerica});
        \item The \textit{ Fonds Boissy}, plays of Louis de Boissy (\url{https://www.licorn-research.fr/Boissy.html});
        \item The \textit{Mercure Galant} project, the famous French \textit{gazette} and literary magazine between 1672 and 1710 (\url{https://obvil.sorbonne-universite.fr/corpus/mercure-galant});
        \item The \textit{Rousseau online} project, works of Jean-Jacques Rousseau (\url{https://www.rousseauonline.ch});
        \item The \textit{Sermo} project, sermons of the 16\textsuperscript{th} and 17\textsuperscript{th}\,c. (\url{http://sermo.unine.ch});
        \item The \textit{Théâtre classique} project, 17\textsuperscript{th} and 18\textsuperscript{th}\,c.~French plays (\url{http://www.theatre-classique.fr});
    \end{itemize}
    \item Additional sources come from researchers who kindly accepted to offer their personal transcriptions or data scrapped by our team:
    \begin{itemize}
        \item Transcriptions of Anne-Élisabeth Spica (17\textsuperscript{th}\,c. French novels);
        \item Transcriptions found on \textit{Wikisource} (\url{https://fr.wikisource.org});
        \item Transcriptions (ePub files) found on \textit{Gallica} (\url{https://gallica.bnf.fr});
        \item Transcriptions found on various websites online.
    \end{itemize}
\end{itemize}

Additional data for later states of the language, up to the 1920's (mainly from FRANTEXT \textit{intégral}), are also provided for two main reasons: on the one hand, it is common to normalise Early Modern French into Contemporary French \cite{gabay_pourquoi_2014} because of the linguistic proximity between these the two states of the language, and on the other hand, it helps to collect (precious) additional data to avoid ending up with with too small of a corpus for our needs.

\begin{table}[htp]
    \centering
    \begin{tabular}{lr}
    \toprule
        Origin & \#Tokens \\
    \midrule
        Spica corpus & 691,467 \\
        Antonomaz project & 119,194 \\
        Acta Pacis Westphlicae II B & 2,463,047 \\
        Bibliothèque Bleue & 776,838 \\
        BVH & 2,434,657 \\
        CEPM & 2,707,432 \\
        Condé project & 3,173,845 \\
        Descartes & 1,025,337 \\
        CELLF & 1,873,772 \\
        Electronic enlightenment & 6,568,047 \\
        Fabula project & 145,978 \\
        \textsc{Frantext} \textit{intégral} ($>$1500, $<$1800) & 60,018,390 \\
        \textsc{Frantext} \textit{intégral} ($>$1800) & 71,504,440 \\
        \textsc{Frantext} \textit{Démonstration} & 1,255,454 \\
        Gallica & 5,212,333 \\
        Boissy project & 438,215 \\
        Mercure galant & 5,427,469 \\
        Rousseau Online project & 2,428,587 \\
        Scrapping & 1,936,835 \\
        Sermo project & 529,647 \\
        Théâtre classique project & 13,916,169 \\
        Wikisource & 996,329 \\
    \midrule
        \textbf{TOTAL} & 185,643,482 \\
    \bottomrule
    \end{tabular}
    \caption{Breakdown of the \freemmax corpus by text origin.}
    \label{tab:my_label}
\end{table}

\begin{figure}[htp]
    \centering
    \includegraphics[width=1\linewidth]{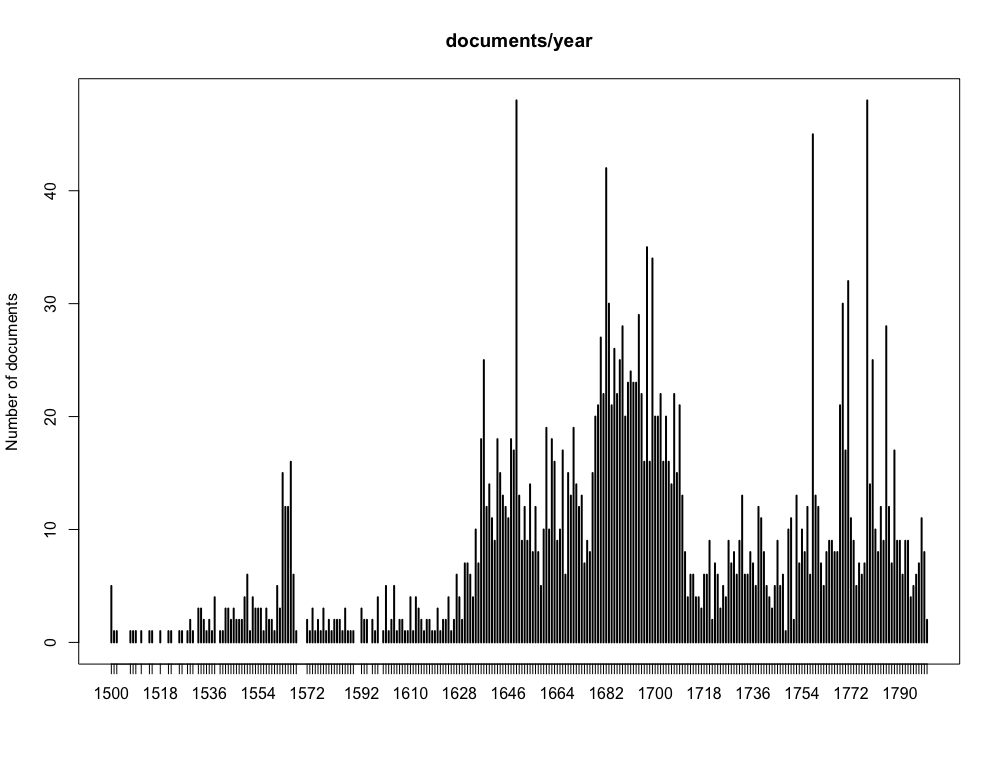}
    \caption{Distribution of the documents in the \freemmax corpus per year}
    \label{fig:FreEMmax_desc}
\end{figure}

The final result is far from being balanced or representative (see Figure~\ref{fig:FreEMmax_desc}). 16\textsuperscript{th}\,c. French documents are under-represented, as well as 18\textsuperscript{th}\,c.~literature. The 17\textsuperscript{th}\,c. is clearly over-represented, especially its second half---probably one of the most important of French literature, which could explain this situation (on top of our personal interest for this specific period).

\begin{figure}[htp]
    \centering
    \includegraphics[width=1\linewidth]{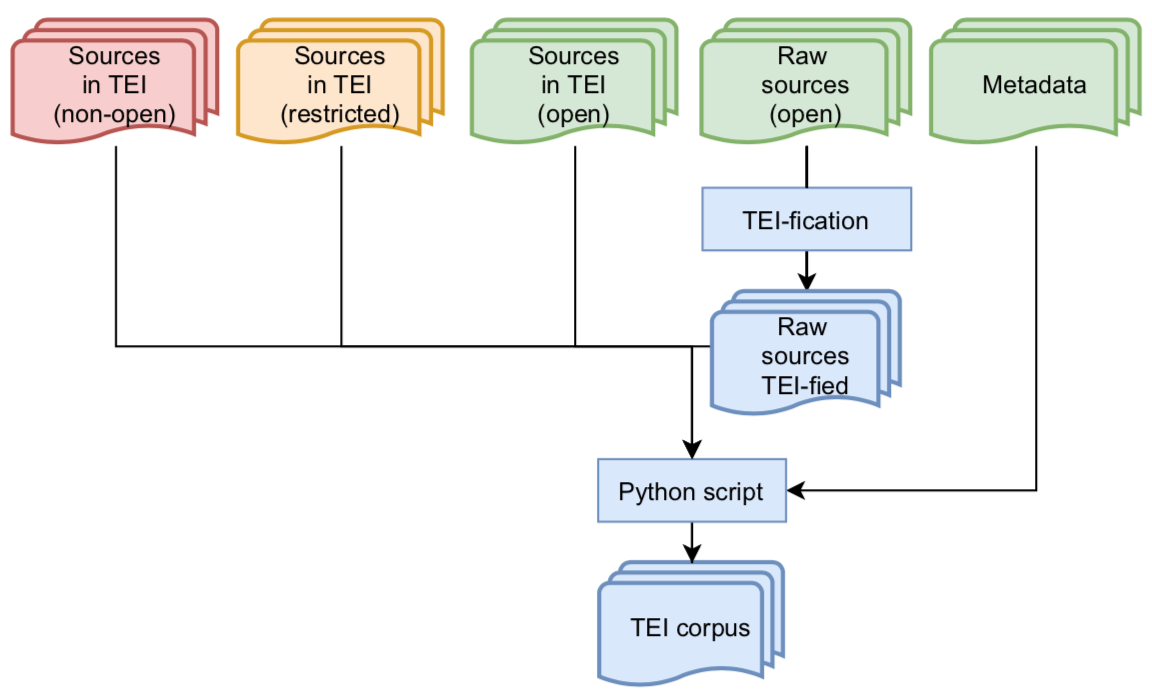}
    \caption{\freemmax compilation pipeline. All files are kept in their original format. Metadata is manually prepared in separate files in order to automatically transform and clean (in blue) all the available documents into XML TEI files following the same encoding. It allows us to distribute open data (in green) but also data distributed with restrictions regarding the modification of the original format (in orange). Non-open texts (in red) are not distributed.}
    \label{fig:pipeline}
\end{figure}

As some texts are still (partially) protected by restrictive licences, the \freemmax corpus exists in both open and non-open versions, only the open one being distributed. In order to limit the impact of licences forbidding the modification of files, we have designed a pipeline to distribute the data as it was found and recreate it (see Figure~\ref{fig:pipeline}).

Metadata is prepared manually in order to have the same categories for each document, whatever its origin. As well as the author, the title and the date (where relevant), we also provide the genre (``theatre''), sometimes a subgenre (``tragedy''), the linguistic status (normalised or not) and the licence attached to the transcription.

\subsection{\texorpdfstring{\freemlpm}{FREEM LPM}}\label{freem_lpm}

The \freemlpm (``Lemma, POS tags, Morphology'') has already been presented \cite{gabay_standardizing_2020}. The POS-annotated data, is a mixture of two different sources. On the one hand, there is the \textit{CornMol} corpus \cite{camps_corpus_2020}, made up of normalised 17\textsuperscript{th}\,c.~French comedies. On the other hand, there is a gold subset of the \textit{Presto} corpus \cite{blumenthal_presto_2017}, made up of texts of different genres written during the 16\textsuperscript{th}, 17\textsuperscript{th} and 18\textsuperscript{th}\,c., which have previously used to train annotation tools \cite{diwersy_ressources_2017}, and was heavily corrected by us to match our annotation principles \cite{gabay_manuel_2020}.

On top of traditional in-domain tests, an out-of-domain testing dataset was prepared to control the capacity of the model to generalise to other genres and periods. Centuries covered are the 16\textsuperscript{th}, 17\textsuperscript{th}, 18\textsuperscript{th}, 19\textsuperscript{th} and 20\textsuperscript{th}. There are two test sets for each century: one made up only of theatre, the other of everything but theatre. Each test set comprises 10 short samples (c.\,100 tokens), as representative as possible of the linguistic production of the century (female and male authors, decade of publication, genre, etc.).

All the data from \freemlpm (but almost none of the out-of-domain) can be found in \freemmax.

\section{D'AlemBERT: a neural language model for Early Modern French}\label{sec:dAlemBERT}
In this section, we describe the pretraining data, architecture, training objective and optimisation setup we use for \dalembert, our new neural language model for Early Modern French.

\subsection{Pre-processing}
Similar to \roberta \cite{liu-etal-2019-roberta} we segment the input text data into subword units using Byte-Pair encoding (BPE) \cite{sennrich-etal-2016-neural} in the implementation proposed by \cite{radford-etal-2019-language} that uses bytes instead of unicode characters as the base subword units. The BPE encoding does not require pre-tokenisation (at the word or token level), thus removing the need to develop a specific tokeniser for Early Modern French. We use a vocabulary size of 32,768 subword tokens. These subwords are learned on the entire \freemmax dataset.

\subsection{Language Modelling}

\begin{table*}[ht]
    \centering\small
    \resizebox{\linewidth}{!}{
        \begin{tabular}{lrrrrrr}
            \toprule
            \multicolumn{7}{c}{\textsc{Original}}                                                      \\
            \midrule
            Model        & 16             & 17             & 18             & 19 & 20 & Avg            \\
            \midrule
            \multicolumn{7}{l}{\hspace*{6mm}\emph{Drama}}                                              \\
            \pieextended & \emph{90.34}   & \emph{94.47}   & \emph{94.64}   & -  & -  & \emph{93.15}   \\
            \camembert   & 87.06          & 89.01          & 90.92          & -  & -  & 89.00          \\
            \dalembert   & \textbf{94.17} & \textbf{96.59} & \textbf{96.28} & -  & -  & \textbf{95.68} \\
            \multicolumn{7}{l}{\hspace*{6mm}\emph{Varia}}                                              \\
            \pieextended & \emph{89.85}   & \emph{93.44}   & \emph{95.98}   & -  & -  & \emph{93.09}   \\
            \camembert   & 86.90          & 88.85          & 92.85          & -  & -  & 89.53          \\
            \dalembert   & \textbf{93.86} & \textbf{95.73} & \textbf{96.95} & -  & -  & \textbf{95.51} \\
            \multicolumn{7}{l}{\hspace*{6mm}\emph{Both}}                                               \\
            \pieextended & \emph{90.08}   & \emph{93.95}   & \emph{95.33}   & -  & -  & \emph{ 93.12}  \\
            \camembert   & 86.98          & 88.93          & 91.89          & -  & -  & 89.27          \\
            \dalembert   & \textbf{94.02} & \textbf{96.16} & \textbf{96.62} & -  & -  & \textbf{95.60} \\
            \bottomrule
        \end{tabular}
        \begin{tabular}{lrrrrrr}
            \toprule
            \multicolumn{7}{c}{\textsc{Normalised or Contemporary}}                                                            \\
            \midrule
            Model        & 16             & 17             & 18             & 19             & 20             & Avg            \\
            \midrule
            \multicolumn{7}{l}{\hspace*{6mm}\emph{Drama}}                                                                      \\
            \pieextended & \emph{93.69}   & \emph{95.75}   & \emph{95.61}   & \emph{95.03}   & \emph{93.71}   & \emph{94.76}   \\
            \camembert   & 90.18          & 91.51          & 91.37          & 91.13          & 91.42          & 91.12          \\
            \dalembert   & \textbf{96.25} & \textbf{96.97} & \textbf{96.80} & \textbf{96.25} & \textbf{95.00} & \textbf{96.25} \\
            \multicolumn{7}{l}{\hspace*{6mm}\emph{Varia}}                                                                      \\
            \pieextended & \emph{92.52}   & \emph{94.81}   & \emph{95.98}   & \emph{92.24}   & \emph{94.03}   & \emph{93.94}   \\
            \camembert   & 89.79          & 90.69          & 93.06          & 90.54          & 89.78          & 93.94          \\
            \dalembert   & \textbf{94.52} & \textbf{96.64} & \textbf{96.88} & \textbf{94.90} & \textbf{95.30} & \textbf{95.65} \\
            \multicolumn{7}{l}{\hspace*{6mm}\emph{Both}}                                                                       \\
            \pieextended & \emph{93.08}   & \emph{95.28}   & \emph{95.80}   & \emph{93.65}   & \emph{93.87}   & \emph{94.35}   \\
            \camembert   & 89.99          & 91.10          & 92.22          & 90.84          & 90.60          & 92.53          \\
            \dalembert   & \textbf{95.39} & \textbf{96.81} & \textbf{96.84} & \textbf{95.58} & \textbf{95.15} & \textbf{95.95} \\
            \bottomrule
        \end{tabular}
    }
    \caption{Comparison between \dalembert, \camembert and \pieextended performance on \freemlpm.}
    \label{tab:POS}
\end{table*}

\paragraph{Transformer}
\dalembert uses the exact same architecture as \roberta, which is a multi-layer bidirectional Transformer \cite{vaswani-etal-2017-attention}.
\dalembert uses the original \emph{base} architecture of \roberta (12 layers, 768 hidden dimensions, 12 attention heads, 110M parameters).

\paragraph{Pretraining Objective}
We train our model on the Masked Language Modelling (MLM) task as proposed by RoBERTa's authors \cite{liu-etal-2019-roberta}: given an input text sequence composed of $N$ tokens $x_1, ..., x_N$, we select 15\% of tokens for possible replacement. Among those selected tokens, 80\% are replaced with the special \texttt{<MASK>} token, 10\% are left unchanged and 10\% are replaced by a random token. The model is then trained to predict the masked tokens using cross-entropy loss.

Again, following the \roberta approach, we dynamically mask tokens instead of fixing them statically for the whole dataset during preprocessing. We also choose not to use the next sentence prediction (NSP) task originally used in \bert \cite{devlin-etal-2019-bert}, as it has been shown that it does not improve downstream task performance \cite{conneau-lample-2019-cross,liu-etal-2019-roberta}.

\paragraph{Optimisation}
We optimise our model in the exact same way as \cite{liu-etal-2019-roberta} using Adam \cite{kingma-ba-2015-adam} ($\beta_1 = 0.9$, $\beta_2 = 0.98$) for 100k steps with large batch sizes of 8,192 sequences, each sequence containing at most 512 tokens.

\paragraph{Pre-training}
We use the \roberta implementation in the Zelda Rose library,\footnote{\url{https://github.com/LoicGrobol/zeldarose}} and again, in the same way as \newcite{liu-etal-2019-roberta} our learning rate is warmed up for 10k steps up to a peak value of $0.0003$ instead of the original $0.0001$ used by the original implementation of \roberta \cite{liu-etal-2019-roberta}, as our model diverged with the $0.0001$ value. We hypothesise that this is either due to the smaller size of \freemmax (compared to the corpora used for \roberta or \camembert) or to our large batch size. We train our model for 31k steps, which amounts to 41 epochs. The total pre-training times, the details of the infrastructure we used and even the carbon emissions of our model are reported in Appendix~\ref{carbon-footprint}.

\section{Evaluation and Discussion}

In order to evaluate our \dalembert model, we fine-tune it for POS tagging on the \freemlpm corpus. We use the \texttt{flair} framework\footnote{\url{https://github.com/flairNLP/flair}} for sequence tagging \cite{akbik-etal-2019-flair}. To fine-tune \dalembert for POS we follow the same approach as \newcite{schweter-akbik-2020-flert} with some modifications: we append a linear layer of size 256 that takes as input the last hidden representation of the \texttt{<s>} special token and the mean of the last hidden representation of the subword units of each token (token as defined for \freemlpm), that is, we use a \emph{``mean''} subword pooling strategy. We fine-tune \dalembert with a learning rate of 0.000005 for a total of 10 epochs. We also fine-tune \camembert using the exact same hyperparameters as that we use for \dalembert.

\freemlpm provides a standard split (train, dev, test), however it also proposes an evaluation on a \emph{out-of-domain} subcorpus that is not contained in the standard split and that is separated by century (from the 16\textsuperscript{th} to the 20\textsuperscript{th} century) and that also contains both the \emph{Normalised} and \emph{Original} versions of the texts for the 16\textsuperscript{th}, 17\textsuperscript{th} and 18\textsuperscript{th} centuries. The idea of this out-of-domain evaluation corpus is to have a fine-grained evaluation of the models to better assess their performance in all the different types of text that one might encounter when working with Early Modern French data.

Following the approach of \newcite{clerice-2020-pie}, we report the scores obtained on the out-of-domain testing dataset of \freemlpm in Table~\ref{tab:POS}. We use the scores previously reported by \newcite{clerice-2020-pie} using \emph{Pie Extended} as our baseline as well as the fine-tuned \camembert that serves as a second baseline as well as a rough estimation of how much knowledge can \dalembert transfer from the \freemmax into this task.

We can see that \dalembert consistently outperforms \pieextended and \camembert in both the normalised and original versions of our out-of-domain testing data and for all different periods by a considerable margin. We can also see that on average the difference in score between \dalembert and \pieextended is greater for the original split than the normalised one. This suggests that \dalembert can generalise more effectively to non-normalised data than the more traditional architecture used by \pieextended. Moreover we can also see that the difference in scores is also greater for the 16\textsuperscript{th}\,c. and 17\textsuperscript{th}\,c. data. This is interesting, especially for the 16\textsuperscript{th}\,c, because, as we can see in Figure~\ref{fig:FreEMmax_desc}, this is the least represented period in the \freemmax corpus. This result actually suggests that \dalembert might be able to do effective transfer learning from the 18\textsuperscript{th}\,c., 19\textsuperscript{th}\,c. and 20\textsuperscript{th}\,c. data to the 16\textsuperscript{th}\,c. and 17\textsuperscript{th}\,c. data.

As for \camembert, we can see that it consistently scores lower than both \dalembert and \pieextended. Moreover, we can see that it struggles particularly with the non-normalised data of the 16\textsuperscript{th}\,c., 17\textsuperscript{th}\,c. and 18\textsuperscript{th}\,c.. This results clearly shows that \camembert cannot easily generalised to these earlier states of languages, or at least not with the quantity of data found in the training set of \freemlpm. These results also show the impressive capacity of \dalembert of quickly generalising to diverse set of states of language, as well as its capacity to transfer knowledge from the \freemmax corpus into this task. The obtained results are also a testament to the importance of the pre-training data, specially taking in account that the pre-training set of \camembert is more than 100 times bigger than that of \dalembert.

\section{Conclusion}

In this paper we presented the manually curated \freemmax corpus of Early Modern French as well as \dalembert, a RoBERTa-based language model trained on \freemmax. With \dalembert, we showed that it is possible to successfully train a transformer-based language model for historical French with even less data than originally shown in previous works \cite{martin-etal-2020-camembert}. Moreover with our POS tagging evaluation we were able to observe some form of transfer learning and generalisation across multiple states of the language corresponding to different periods of time. Both our corpus and our model will be of use to digital humanists and linguists interested in Early Modern French. For our future work, we hope that will be able to study the application of our \dalembert model to other NLP tasks such as text normalisation, named entity recognition and even document structuring, where we hope to more extensively study the transfer learning capabilities of our approach.

\section{Acknowledgements}
We would like to warmly thank Karine Abiven, Bertrand Gaiffe, Annette Gerstenberg, Pierre Larrivée, Gaël Lejeune, Laurent Romary, Anne-Élisabeth Spica and Martin Wynne for their help in gathering the data. This work was also performed using HPC resources from GENCI-IDRIS (Grant 2021-AD011011330R1).

This work was also partly funded by Rachel Bawden's and Benoît Sagot's chairs in the PRAIRIE institute funded by the French national agency ANR as part of the ``Investissements d'avenir'' programme under the reference ANR-19-P3IA-0001, as well as the BASNUM ANR project (ANR-18-CE38-0003).

% \nocite{*}
\section{Bibliographical References}\label{reference}
%\label{main:ref}

\bibliographystyle{lrec2022-bib}
\bibliography{bibliography,anthology}

\begin{thebibliography}{}

\bibitem[\protect\citename{{ATILF}}1998  a]{atilf_frantext_1998-1}
{ATILF}.
\newblock (1998--a).
\newblock {\em Frantext Démonstration}.
\newblock {ATILF} - {CNRS} \& Université de Lorraine.

\bibitem[\protect\citename{{ATILF}}1998  b]{atilf_frantext_1998}
{ATILF}.
\newblock (1998--b).
\newblock {\em Frantext intégral}.
\newblock {ATILF} - {CNRS} \& Université de Lorraine.

\bibitem[\protect\citename{Blumenthal and Vigier}2018]{presto_corpus_2018}
Blumenthal, Peter and Vigier, Denis (dir.).
\newblock (2018).
\newblock {\em Presto: corpus noyau}.

\bibitem[\protect\citename{{Bodleian Libraries}}2008
  ]{bodleian_libraries_electronic_2008}
{Bodleian Libraries}.
\newblock (2008--).
\newblock {\em Electronic Enlightenment}.
\newblock Oxford University Press.

\bibitem[\protect\citename{Camps \bgroup et al.\egroup }2019]{camps_geste_2016}
Camps, Jean-Baptiste and {LAKME-ENC} and Cochet, Alice and Ing, Lucence and
  Paulinelvq.
\newblock (2019).
\newblock {\em {Jean-Baptiste-Camps/Geste: Geste: un corpus de chansons de
  geste, 2016-…}}.
\newblock Zenodo.

\bibitem[\protect\citename{Demonet}1998  ]{demonet_epistemon_1998}
Demonet, Marie-Luce (dir.).
\newblock (1998--).
\newblock {\em Epistemon}.
\newblock Centre d’Etudes Supérieures de la Renaissance.

\bibitem[\protect\citename{Erjavec}2015]{erjavec_reference_2015}
Erjavec, Tomaž.
\newblock (2015).
\newblock {\em Reference corpus of historical Slovene goo300k 1.2}.

\bibitem[\protect\citename{Gabay \bgroup et al.\egroup }2022]{FREEMmax}
Gabay, Simon and Bartz, Alexandre and Gambette, Philippe and Chagué, Alix.
\newblock (2022).
\newblock {\em FreEM max OA: A Large Corpus for Early modern French - Open
  access version}.
\newblock Université de Genève, 1.0.

\bibitem[\protect\citename{Morrissey and Olsen}1981  ]{morrissey_artfl_1981}
Morrissey, Robert and Olsen, Mark.
\newblock (1981--).
\newblock {\em American and French Research on the Treasury of the French
  Language (ARTFL)}.
\newblock University of Chicago.

\bibitem[\protect\citename{{Reenen, Pieter van and Mulder,
  Maaike}}1998]{Reenen_corpus_1998}
{Reenen, Pieter van and Mulder, Maaike}.
\newblock (1998).
\newblock {\em Corpus Middelnederlands}.
\newblock Instituut voor de Nederlandse Taal, 1.0.

\end{thebibliography}


\begin{thebibliography}{}

\bibitem[\protect\citename{Akbik \bgroup et al.\egroup
  }2019]{akbik-etal-2019-flair}
Akbik, A., Bergmann, T., Blythe, D., Rasul, K., Schweter, S., and Vollgraf, R.
\newblock (2019).
\newblock {FLAIR}: An easy-to-use framework for state-of-the-art {NLP}.
\newblock In {\em Proceedings of the 2019 Conference of the North {A}merican
  Chapter of the Association for Computational Linguistics (Demonstrations)},
  pages 54--59, Minneapolis, Minnesota, June. Association for Computational
  Linguistics.

\bibitem[\protect\citename{Amatuzzi \bgroup et al.\egroup
  }2019]{amatuzzi_ameliorer_2019}
Amatuzzi, A., Skupien~Dekens, C., Ayres-Bennett, W., Gerstenberg, A., and
  Schoesler, L.
\newblock (2019).
\newblock Améliorer et appliquer les outils numériques. ressources et
  approches pour l'étude du changement linguistique en français préclassique
  et classique.
\newblock In {\em Le français en Diachronie}, Travaux de Linguistique Romane,
  pages 337--364. Editions de linguistique et de philologie.

\bibitem[\protect\citename{Ayres-Bennett and
  Seijido}2011]{ayres-bennett_remarques_2011}
Wendy Ayres-Bennett et~al., editors.
\newblock (2011).
\newblock {\em Remarques et observations sur la langue française. Histoire et
  évolution d’un genre}.
\newblock Number~1 in Histoire et évolution du français. Classiques Garnier.

\bibitem[\protect\citename{Bamman and Burns}2020]{Bamman_latinBErt_2020}
Bamman, D. and Burns, P.~J.
\newblock (2020).
\newblock Latin {BERT:} {A} contextual language model for classical philology.
\newblock {\em CoRR}, abs/2009.10053.

\bibitem[\protect\citename{Bender \bgroup et al.\egroup
  }2021]{bender-etal-2021-on}
Bender, E.~M., Gebru, T., McMillan-Major, A., and Shmitchell, S.
\newblock (2021).
\newblock On the dangers of stochastic parrots: Can language models be too big?
\newblock In {\em Proceedings of the 2021 ACM Conference on Fairness,
  Accountability, and Transparency}, FAccT '21, page 610–623, New York, NY,
  USA. Association for Computing Machinery.

\bibitem[\protect\citename{Blumenthal \bgroup et al.\egroup
  }2017]{blumenthal_presto_2017}
Blumenthal, P., Diwersy, S., Falaise, A., Lay, M.-H., Souvay, G., and Vigier,
  D.
\newblock (2017).
\newblock Presto, un corpus diachronique pour le français des {XVIe}-{XXe}
  siècles.
\newblock In {\em Actes de la 24ème conférence sur le Traitement Automatique
  des Langues Naturelles - {TALN}'17}. Association pour le traitement
  automatique des langues.

\bibitem[\protect\citename{Camps \bgroup et al.\egroup
  }2020]{camps_corpus_2020}
Camps, J.-B., Gabay, S., Fièvre, P., Clérice, T., and Cafiero, F.
\newblock (2020).
\newblock {Corpus and Models for Lemmatisation and {POS}-tagging of Classical
  French Theatre}.
\newblock {\em Journal of Data Mining \& Digital Humanities}.

\bibitem[\protect\citename{Clérice}2020]{clerice-2020-pie}
Clérice, T.
\newblock (2020).
\newblock Pie extended, an extension for pie with pre-processing and
  post-processing, June.

\bibitem[\protect\citename{Conneau and Lample}2019]{conneau-lample-2019-cross}
Conneau, A. and Lample, G.
\newblock (2019).
\newblock Cross-lingual language model pretraining.
\newblock In H.~Wallach, et~al., editors, {\em Advances in Neural Information
  Processing Systems}, volume~32. Curran Associates, Inc.

\bibitem[\protect\citename{Desrochers \bgroup et al.\egroup
  }2016]{desrochers-etal-2016-a}
Desrochers, S., Paradis, C., and Weaver, V.~M.
\newblock (2016).
\newblock A validation of dram rapl power measurements.
\newblock In {\em Proceedings of the Second International Symposium on Memory
  Systems}, MEMSYS '16, page 455–470, New York, NY, USA. Association for
  Computing Machinery.

\bibitem[\protect\citename{Devlin \bgroup et al.\egroup
  }2019]{devlin-etal-2019-bert}
Devlin, J., Chang, M.-W., Lee, K., and Toutanova, K.
\newblock (2019).
\newblock {BERT}: Pre-training of deep bidirectional transformers for language
  understanding.
\newblock In {\em Proceedings of the 2019 Conference of the North {A}merican
  Chapter of the Association for Computational Linguistics: Human Language
  Technologies, Volume 1 (Long and Short Papers)}, pages 4171--4186,
  Minneapolis, Minnesota, June. Association for Computational Linguistics.

\bibitem[\protect\citename{Diwersy \bgroup et al.\egroup
  }2017]{diwersy_ressources_2017}
Diwersy, S., Falaise, A., Lay, M.-H., and Souvay, G.
\newblock (2017).
\newblock Ressources et méthodes pour l’analyse diachronique.
\newblock {\em Langages}, N° 206(2):21--44.

\bibitem[\protect\citename{Gabay \bgroup et al.\egroup
  }2020]{gabay_manuel_2020}
Gabay, S., Camps, J.-B., and Clérice, T.
\newblock (2020).
\newblock Manuel d'annotation linguistique pour le français moderne ({XVIe}
  -{XVIIIe} siècles).

\bibitem[\protect\citename{Gabay \bgroup et al.\egroup }2020
  10]{gabay_standardizing_2020}
Gabay, S., Clérice, T., Camps, J.-B., Tanguy, J.-B., and Gille-Levenson, M.
\newblock (2020-10).
\newblock {Standardizing linguistic data: method and tools for annotating
  (pre-orthographic) French}.
\newblock In {\em Proceedings of the 2nd International Digital Tools \& Uses
  Congress ({DTUC} '20)}, Hammamet, Tunisia.

\bibitem[\protect\citename{Gabay}2014]{gabay_pourquoi_2014}
Gabay, S.
\newblock (2014).
\newblock Pourquoi moderniser l’orthographe? principes d’ecdotique et
  littérature du {XVIIe} siècle.
\newblock {\em Vox Romanica}, 73(1):27--42.

\bibitem[\protect\citename{Guez~de Balzac}1624]{guez_de_balzac_lettres_1624}
Guez~de Balzac, J.-L.
\newblock (1624).
\newblock {\em Lettres du sieur de Balzac}.
\newblock T. Du Bray.

\bibitem[\protect\citename{Kingma and Ba}2015]{kingma-ba-2015-adam}
Kingma, D.~P. and Ba, J.
\newblock (2015).
\newblock Adam: {A} method for stochastic optimization.
\newblock In Yoshua Bengio et~al., editors, {\em 3rd International Conference
  on Learning Representations, {ICLR} 2015, San Diego, CA, USA, May 7-9, 2015,
  Conference Track Proceedings}.

\bibitem[\protect\citename{Le \bgroup et al.\egroup
  }2020]{le-etal-2020-flaubert}
Le, H., Vial, L., Frej, J., Segonne, V., Coavoux, M., Lecouteux, B., Allauzen,
  A., Crabb{\'e}, B., Besacier, L., and Schwab, D.
\newblock (2020).
\newblock {F}lau{BERT} : des mod{\`e}les de langue contextualis{\'e}s
  pr{\'e}-entra{\^\i}n{\'e}s pour le fran{\c{c}}ais ({F}lau{BERT} :
  Unsupervised language model pre-training for {F}rench).
\newblock In {\em Actes de la 6e conf{\'e}rence conjointe Journ{\'e}es
  d'{\'E}tudes sur la Parole (JEP, 33e {\'e}dition), Traitement Automatique des
  Langues Naturelles (TALN, 27e {\'e}dition), Rencontre des {\'E}tudiants
  Chercheurs en Informatique pour le Traitement Automatique des Langues
  (R{\'E}CITAL, 22e {\'e}dition). Volume 2 : Traitement Automatique des Langues
  Naturelles}, pages 268--278, Nancy, France, 6. ATALA et AFCP.

\bibitem[\protect\citename{{Liu} \bgroup et al.\egroup
  }2019]{liu-etal-2019-roberta}
{Liu}, Y., {Ott}, M., {Goyal}, N., {Du}, J., {Joshi}, M., {Chen}, D., {Levy},
  O., {Lewis}, M., {Zettlemoyer}, L., and {Stoyanov}, V.
\newblock (2019).
\newblock {RoBERTa: A Robustly Optimized BERT Pretraining Approach}.
\newblock {\em arXiv e-prints}, page arXiv:1907.11692, July.

\bibitem[\protect\citename{Manjavacas \bgroup et al.\egroup
  }2019]{manjavacas-etal-2019-improving}
Manjavacas, E., K{\'a}d{\'a}r, {\'A}., and Kestemont, M.
\newblock (2019).
\newblock Improving lemmatization of non-standard languages with joint
  learning.
\newblock In {\em Proceedings of the 2019 Conference of the North {A}merican
  Chapter of the Association for Computational Linguistics: Human Language
  Technologies, Volume 1 (Long and Short Papers)}, pages 1493--1503,
  Minneapolis, Minnesota, June. Association for Computational Linguistics.

\bibitem[\protect\citename{Martin \bgroup et al.\egroup
  }2020]{martin-etal-2020-camembert}
Martin, L., Muller, B., Ortiz~Su{\'a}rez, P.~J., Dupont, Y., Romary, L., de~la
  Clergerie, {\'E}., Seddah, D., and Sagot, B.
\newblock (2020).
\newblock {C}amem{BERT}: a tasty {F}rench language model.
\newblock In {\em Proceedings of the 58th Annual Meeting of the Association for
  Computational Linguistics}, pages 7203--7219, Online, July. Association for
  Computational Linguistics.

\bibitem[\protect\citename{{Martin, Robert
  (dir.)}}2020]{martin_dictionnaire_2020}
{Martin, Robert (dir.)}.
\newblock (2020).
\newblock {\em Dictionnaire du Moyen Français}.
\newblock {ATILF} - {CNRS} \& Université de Lorraine.

\bibitem[\protect\citename{Moretti}2013]{moretti_distant_2013}
Moretti, F.
\newblock (2013).
\newblock {\em Distant reading}.
\newblock Verso.

\bibitem[\protect\citename{{Ortiz Su{\'a}rez} \bgroup et al.\egroup
  }2019]{ortiz-suarez-etal-2019-asynchronous}
{Ortiz Su{\'a}rez}, P.~J., Sagot, B., and Romary, L.
\newblock (2019).
\newblock Asynchronous pipelines for processing huge corpora on medium to low
  resource infrastructures.
\newblock In Piotr Bański, et~al., editors, {\em Proceedings of the Workshop
  on Challenges in the Management of Large Corpora (CMLC-7) 2019. Cardiff, 22nd
  July 2019}, pages 9--16, Mannheim. Leibniz-Institut f{\"u}r Deutsche Sprache.

\bibitem[\protect\citename{Ortiz~Su{\'a}rez \bgroup et al.\egroup
  }2020]{ortiz-suarez-etal-2020-monolingual}
Ortiz~Su{\'a}rez, P.~J., Romary, L., and Sagot, B.
\newblock (2020).
\newblock A monolingual approach to contextualized word embeddings for
  mid-resource languages.
\newblock In {\em Proceedings of the 58th Annual Meeting of the Association for
  Computational Linguistics}, pages 1703--1714, Online, July. Association for
  Computational Linguistics.

\bibitem[\protect\citename{Radford \bgroup et al.\egroup
  }2019]{radford-etal-2019-language}
Radford, A., Wu, J., Child, R., Luan, D., Amodei, D., and Sutskever, I.
\newblock (2019).
\newblock Language models are unsupervised multitask learners.
\newblock {\em OpenAI Blog}, 1:8.

\bibitem[\protect\citename{Rijhwani \bgroup et al.\egroup
  }2021]{rijhwani_lexically_2021}
Rijhwani, S., Rosenblum, D., Anastasopoulos, A., and Neubig, G.
\newblock (2021).
\newblock Lexically aware semi-supervised learning for {OCR} post-correction.
\newblock {\em Transactions of the Association for Computational Linguistics},
  9:1285--1302.

\bibitem[\protect\citename{Schwartz \bgroup et al.\egroup
  }2020]{schwartz-etal-2020-green}
Schwartz, R., Dodge, J., Smith, N.~A., and Etzioni, O.
\newblock (2020).
\newblock Green ai.
\newblock {\em Commun. ACM}, 63(12):54–63.

\bibitem[\protect\citename{{Schweter} and
  {Akbik}}2020]{schweter-akbik-2020-flert}
{Schweter}, S. and {Akbik}, A.
\newblock (2020).
\newblock {FLERT: Document-Level Features for Named Entity Recognition}.
\newblock {\em arXiv e-prints}, page arXiv:2011.06993, November.

\bibitem[\protect\citename{Sennrich \bgroup et al.\egroup
  }2016]{sennrich-etal-2016-neural}
Sennrich, R., Haddow, B., and Birch, A.
\newblock (2016).
\newblock Neural machine translation of rare words with subword units.
\newblock In {\em Proceedings of the 54th Annual Meeting of the Association for
  Computational Linguistics (Volume 1: Long Papers)}, pages 1715--1725, Berlin,
  Germany, August. Association for Computational Linguistics.

\bibitem[\protect\citename{Strubell \bgroup et al.\egroup
  }2019]{strubell-etal-2019-energy}
Strubell, E., Ganesh, A., and McCallum, A.
\newblock (2019).
\newblock Energy and policy considerations for deep learning in {NLP}.
\newblock In {\em Proceedings of the 57th Annual Meeting of the Association for
  Computational Linguistics}, pages 3645--3650, Florence, Italy, July.
  Association for Computational Linguistics.

\bibitem[\protect\citename{Vachon}2010]{vachon_changement_2010}
Vachon, C.~H.
\newblock (2010).
\newblock {\em Le Changement linguistique au {XVIe} siècle: une étude basée
  sur des textes littéraires français}.
\newblock {ELiPhi}, Éditions de linguistique et de philologie.

\bibitem[\protect\citename{Vaswani \bgroup et al.\egroup
  }2017]{vaswani-etal-2017-attention}
Vaswani, A., Shazeer, N., Parmar, N., Uszkoreit, J., Jones, L., Gomez, A.~N.,
  Kaiser, L.~u., and Polosukhin, I.
\newblock (2017).
\newblock Attention is all you need.
\newblock In I.~Guyon, et~al., editors, {\em Advances in Neural Information
  Processing Systems}, volume~30. Curran Associates, Inc.

\end{thebibliography}

\section{Language Resource References}
\label{lr:ref}
\bibliographystylelanguageresource{lrec2022-bib}
\bibliographylanguageresource{languageresource.bib}

\appendix

\section{Carbon Footprint}\label{carbon-footprint}

\begin{table}[th]
    \centering\small
    \resizebox{\linewidth}{!}{
    \begin{tabular}{@{}lrrrr@{}}
        \toprule
        \textbf{Model}                                        & {\textbf{Power (W)}}  & {\textbf{Time (h)}}  & {\textbf{(PUE$\cdotp$kWh)}} & {\textbf{CO\textsuperscript{2}e (kg)}} \\
        \midrule
        Pre-train       & 48640 &  20 & 1537.02 & 46.11 \\
        Evaluation     &  589  & 1 &   0.93 &   0.03 \\
        \midrule
        Total CO\textsuperscript{2}e &       &    &          & 46.14\\
        \bottomrule
    \end{tabular}
    }
    \caption{Average power draw, number of models trained, training times in hours, mean power consumption including power usage effectiveness (PUE), and CO\textsuperscript{2} emissions; for each setting.}
    \label{tab:carbon}
\end{table}

In light of recent interest concerning the energy consumption and carbon emission of machine learning models and specifically of those of language models \cite{schwartz-etal-2020-green,bender-etal-2021-on}, we have decided to report the power consumption and carbon footprint of all our experiments following the approach of \newcite{strubell-etal-2019-energy}. We report the energy consumption and carbon emissions of both the pre-training of D'AlemBERT and its evaluation. 

\paragraph{Pre-training:} We use a cluster of 32 machines, each one having 4 GPU Nvidia Tesla V100 SXM2 32GiB, 192GiB of RAM, and two Intel Xeon Gold 6248 processors. One Nvidia Tesla V100 card is rated at around 300W,\footnote{\href{https://www.nvidia.com/en-us/data-center/v100/}{ Nvidia Tesla V100 specification}} while the Xeon Gold 6248 processor is rated at 150W.\footnote{\href{https://ark.intel.com/content/www/us/en/ark/products/192446/intel-xeon-gold-6248-processor-27-5m-cache-2-50-ghz.html}{Intel Xeon Gold 6248 specification}} For the DRAM we can use the work of \newcite{desrochers-etal-2016-a} to estimate the total power draw of 192GiB of RAM at around 20W. Thus, the total power draw of the pre-training adds up to around 48640W.

\paragraph{Evaluation:} We use a single machine with a single GPU Nvidia Tesla V100 SXM2 32GiB, 384GiB of RAM and two Intel Xeon Gold 6226 processors. The Xeon Gold 6226 processor is rated at 125 W,\footnote{\href{https://ark.intel.com/content/www/us/en/ark/products/193957/intel-xeon-gold-6226-processor-19-25m-cache-2-70-ghz.html}{Intel Xeon Gold 6226 specification}} and the DRAM total power draw can be estimated at around 39W. Therefore, the total power draw of the evaluation adds up to around 589W.

With this information, we use the formula proposed by \newcite{strubell-etal-2019-energy} to compute the total power required for each setting:

\begin{equation*}
    p_t = \frac{1.58t(cp_{c} + p_r + gp_g)}{1000}
\end{equation*}

Where $c$ and $g$ are the number of CPUs and GPUs respectively, $p_c$ is the average power draw (in W) from all CPU sockets, $p_r$ the average power draw from all DRAM sockets and $p_g$ the average power draw of a single GPU. We estimate the total power consumption by adding GPU, CPU and DRAM consumption, and then multiplying by the \emph{Power Usage Effectiveness} (PUE), which accounts for the additional energy required to support the compute infrastructure. We use a PUE coefficient of 1.58, the 2018 global average for data centres \cite{strubell-etal-2019-energy}. In Table~\ref{tab:carbon} we report the training times in hours, as well as the total power draw (in Watts) of the system used to train the models. We use this information to compute the total power consumption of each setting, also reported in Table~\ref{tab:carbon}.

We can further estimate the CO\textsuperscript{2} emissions in kilograms of each single model by multiplying the total power consumption by the average CO\textsuperscript{2} emissions per kWh in our region, which were around 30g/kWh between the 30\textsuperscript{th} and the 31\textsuperscript{st} of December,\footnote{\href{https://www.rte-france.com/eco2mix/les-emissions-de-co2-par-kwh-produit-en-france}{Rte - éCO\textsuperscript{2}mix}.} when the models were trained. Thus the total CO\textsuperscript{2} emissions in kg for one single model can be computed as:
 
\begin{equation*}
    \text{CO}_{2}\text{e} = 0.030 p_t
\end{equation*}

All emissions are also reported in Table~\ref{tab:carbon}.

\end{document}